\documentclass{article}

\usepackage{iclr2023_conference}
\usepackage[utf8]{inputenc}

\usepackage[utf8]{inputenc} 
\usepackage[T1]{fontenc}    
\usepackage{hyperref}       
\usepackage{url}            
\usepackage{booktabs}       
\usepackage{amsfonts}       
\usepackage{nicefrac}       
\usepackage{microtype}      
\usepackage{xcolor}         
\usepackage{amsmath}
\usepackage{hyperref}
\usepackage{array}
\usepackage{algorithm}
\usepackage{amsfonts,amsmath, amssymb, bm}
\usepackage{natbib}
\usepackage{graphicx}
\usepackage{epstopdf}
\usepackage{float}
\usepackage{caption}
\usepackage{subcaption}
\usepackage{times}
\usepackage{sidecap}
\usepackage{multirow}
\usepackage{amsfonts}
\usepackage{algorithmicx,algpseudocode}

\usepackage{array}
\newcolumntype{P}[1]{>{\centering\arraybackslash}p{#1}}

\title{AutoSparse: Towards Automated Sparse Training of Deep neural Networks}
\author{%
  Abhisek Kundu \\
  Parallel Computing Lab, Intel Labs\\
  \texttt{abhisekkundu@gmail.com}
  \And Naveen K. Mellempudi \\
  Parallel Computing Lab, Intel Labs\\
  \texttt{naveen.k.mellempudi@intel.com}
  \And Dharma Teja Vooturi \\
  Parallel Computing Lab, Intel Labs\\
  \texttt{dharma.teja.vooturi@intel.com}
  \And Bharat Kaul \\
  Parallel Computing Lab, Intel Labs\\
  \texttt{bharat.kaul@intel.com}
  \And Pradeep Dubey \\
  Parallel Computing Lab, Intel Labs\\
  \texttt{pradeep.dubey@intel.com}
}
\iclrfinalcopy 
\begin{document}

\maketitle
\begin{abstract}

Sparse training is emerging as a promising avenue for reducing the computational cost of training neural networks. Several recent studies have proposed pruning methods using learnable thresholds to efficiently explore the non-uniform distribution of sparsity inherent within the models. 
%
%
In this paper, we propose Gradient Annealing (GA), 
where gradients of masked weights are scaled down in a non-linear manner. GA provides an elegant trade-off between sparsity and accuracy without the need for additional sparsity-inducing regularization.
We integrated GA with the latest learnable pruning methods to create an automated sparse training algorithm called \textit{AutoSparse}, which achieves better accuracy and/or training/inference FLOPS reduction than existing learnable pruning methods for sparse ResNet50 and MobileNetV1 on ImageNet-1K: AutoSparse achieves ($2\times$, $7\times$)    reduction in (training,inference) FLOPS for ResNet50 on ImageNet at $80\%$ sparsity.
%
%
%
Finally, AutoSparse outperforms sparse-to-sparse SotA method MEST (uniform sparsity) for $80\%$ sparse ResNet50 with similar accuracy, where MEST uses $12\%$ more training FLOPS and $50\%$ more inference FLOPS.

\end{abstract}
\section{Introduction}
Deep learning models (DNNs) have emerged as the preferred solution for many important problems in the domains of computer vision, 
language modeling, 
recommendation systems 
and reinforcement learning. 
Models have grown larger and more complex over the years, as they are applied to increasingly difficult problems on ever-growing datasets. In addition, 
DNNs are designed to operate in overparameterized regime \cite{arora_overparam, bias_variance,empirical_overparam} to facilitate easier optimization using gradient descent methods. 
Consequently, computational costs (memory and floating point operations FLOPS) of performing training and inference tasks on state-of-the-art (SotA) models has been growing at an exponential rate (\cite{exp_compute}). 

Two major techniques to make DNNs more efficient are (1) reduced-precision 
\cite{fp16_mixed_nv, ibm_fp8_1, ibm_fp8_2,int16_intel, intel_bf16,mellempudi2020mixed}, 
and (2) sparse representation. 
Today, SotA training hardware consists of significantly more reduced-precision FLOPS compared to traditional FP32 computations, while support for structured sparsity is also emerging \cite{2_4sparsity2021}.
%
In sparse representation, selected model parameters are masked/pruned resulting in significant reduction in FLOPS, which has a potential of more than $10\times$ throughput. 
While overparameterized models help to ease the training, only a fraction of the model parameters is needed to perform inference accurately \cite{speech_sparse, Han2015,rnn_sparsity,state-sparsity2019,train_large_compress}. Modern DNNs have large number of layers with non-uniform distribution of number of model parameters as well as FLOPS per layer. 
Sparse training algorithms aim to achieve an overall sparsity budget using layer-wise (a) uniform sparsity (b) heuristic non-uniform sparsity, or (c) learned non-uniform sparsity. First two techniques can cause sub-optimal parameter allocation which leads to drop in accuracy and/or lower FLOPS efficiency.   

Sparse training methods can be categorized based on when and how to prune. Producing sparse model from fully-trained dense model (\cite{Han2015}) is useful for efficient inference and not for efficient training. 
Sparse-to-sparse methods MEST (\cite{MEST2021}), despite being efficient in each iteration, requires longer training epochs (reducing the gain in training FLOPS) or frequent dense gradient update TopKAST (\cite{Top-KAST2021}) (more FLOPS during backpass) while keeping first/last layer dense to achieve high accuracy results. 
Dense-to-sparse methods monotonically increase sparsity from a dense, untrained model throughout training GMP (\cite{GMP2017}) using some schedule, and/or use full gradient updates to masked weights OptG (\cite{OptG2022}) to improve accuracy. These methods typically have higher training FLOPS count.

\textbf{Learning Sparsity-Accuracy Trade-off: }
Recent methods that learn sparsity distribution during training are STR (\cite{STR2020}), DST (\cite{DST2020}), LTP (\cite{LTP2021}), CS (\cite{CS2021}) and SCL (\cite{SCL2022}). Learned sparsity methods offer a two-fold advantage over methods with uniform sparsity and heuristically allocated sparsity: (1) computationally efficient, as the overhead of pruning mask computation (e.g. choosing top $k$ largest) is eliminated, (2) learns a non-uniform sparsity distribution inherent within the model, producing a more FLOPS-efficient sparse model for inference. 
However, the main challenge with these methods is to identify the sparsity distribution in early epochs, in order to become competitive to sparse-to-sparse methods in terms of training FLOPS, while achieving sparsity-accuracy trade-off.
For this, these methods typically apply sparsity-inducing regularizer with the objective function (except STR). However, these methods either produce sub-optimal sparsity/FLOPS (SCL) or suffer from significant accuracy loss (LTP), or susceptible to `run-away sparsity' while extracting high sparsity from early iterations  (STR, DST). In order to deal with uncontrolled sparsity growth, DST implements hard upper limit checks (e.g. $99\%$) on sparsity to trigger a reset of the offending threshold (falling into dense training regime and resulting in higher training FLOPS count) to prevent loss of accuracy. Similarly, STR uses a combination of small initial thresholds  with an appropriate weight decay to delay the induction of sparsity until later in the training cycle (e.g. $30$ epochs) to control unfettered growth of sparsity. 

Motivated by the above challenges, we propose \textbf{Gradient Annealing (GA)} method, to address the aforementioned issues related to training sparse models. Comparing with existing methods, \textbf{GA} offers greater flexibility to explore the trade-off between model sparsity and accuracy, and  provides greater stability by preventing divergence due to runaway sparsity. We also propose a unified training algorithm called \textbf{AutoSparse}, combining best of learnable threshold methods STR with \textbf{GA} that attempts to pave the path towards full automation of sparse-to-sparse training. Additionally, we also demonstrated that when coupled with uniform pruning methods TopKAST, \textbf{GA} can extract better accuracy at a $80\%$ sparsity budget.
%
The key contributions of this work are as follows:

$\bullet$
We present a novel \textit{Gradient Annealing} (GA) method (with hyper-parameter $\alpha$) which is a more generalized gradient approximator than STE (\cite{STE2013}) and ReLU.
For training end-to-end sparse models, GA provides greater flexibility for sparsity-accuracy trade-off than other methods.

 $\bullet$
 We propose \textit{AutoSparse}, an algorithm that combines GA with learnable sparsity method STR to create an automated framework for end-to-end sparse training. AutoSparse outperforms SotA learnable sparsity methods in terms of accuracy and training/inference FLOPS 
 by maintaining consistently high sparsity throughout the training  
 for ResNet50 and MobileNetV1 on Imagenet-1K.
 %
 %
 
 %
 $\bullet$ 
 We demonstrate the efficacy of \textit{Gradient Annealing} as a general learning technique independent of \textit{AutoSparse} by applying it to \textit{TopKAST} method to improve the Top-1 accuracy of ResNet50 by $0.3\%$ on ImageNet-1K dataset using uniform sparsity budget of $80\%$.
\\\\
$\bullet$ 
Finally, \textit{AutoSparse} outperforms SotA sparse-to-sparse training method MEST (uniform sparsity) where MEST uses (12\%, 50\%) more FLOPS for (training,inference) than AutoSparse for 80\% sparse ResNet50 with comparable accuracy.

\section{Related Work}
Sparse training methods can be categorized the following way based on when and how to prune.
\\
\textbf{Pruning after Training: }
Here sparse models are created using a three-stage pipeline-- train a dense model, prune, and re-train (\cite{Han2015, DNS2016}). These methods are computationally more expensive than dense training and are useful only for FLOPS and memory reduction in inference.
\\
\textbf{Pruning before Training: }
Such methods are inspired by the `Lottery Ticket Hypothesis' (\cite{lottery_ticket}), which 
tries to find a sparse weight mask (subnetwork) that can be trained from scratch to match the quality of dense trained model. However, they use an iterative pruning scheme that is repeated for several full training passes to find the mask. SNIP (\cite{SNIP2019}) preserves the loss after pruning based on connection sensitivity. GraSP (\cite{Grasp2020}) prunes connection such that it preserves the network's gradient flow. SynFlow (\cite{SynFlow2020}) uses iterative synaptic flow pruning to preserve the total flow of synaptic strength (global pruning at initialization with no data and no backpropagation). 3SP (\cite{3SP2020}) introduces compute-aware pruning of channels (structured pruning). The main advantage of these methods is that they are compute and memory efficient in every iteration. However, they often suffer from significant loss of accuracy. 
Note that, the weight pruning mask is fixed throughout the training. 
\\
\textbf{Pruning during Training: }
For these methods, the weight pruning mask evolves dynamically with training. Methods of this category can belong to either sparse-to-sparse or dense-to-sparse training. 
In \textit{sparse-to-sparse training}, we have a sparse model to start with (based on sparsity budget) and the budget sparsity is maintained throughout the training. 
SET (\cite{SET2018}) pioneered this approach where they replaced a fraction of least magnitude weights by random weights for better exploration. 
DSR (\cite{DSR2019}) allowed sparsity budget to be non-uniform across layers heuristically, e.g., higher sparsity for later layers. 
SNFS (\cite{SNFS2019}) proposed to use momentum of each parameter as a criterion to grow the weights leading to increased accuracy. However, this method requires computing full gradient for all the weights even for masked weights. 
RigL (\cite{Rigl2021}) activates/revives new weights ranked according to gradient magnitude (using infrequent full gradient calculation), i.e., masked weights receive gradients only after certain iterations. They decay this number of revived weights with time. The sparsity distribution can be uniform or non-uniform (using ERK where sparsity is scaled using number of neurons and kernel dimension).  
MEST (\cite{MEST2021}) always maintains fixed sparsity in forward and backward pass by computing gradients of survived weights only. For better exploration, they remove some of the least important weights (ranked proportional to magnitude plus the gradient magnitude) and introduce same number of random ‘zero’ weights to maintain the sparsity budget. This method is suitable for sparse training on edge devices with limited memory.
Restricting gradient flow causes accuracy loss in sparse-to-sparse methods despite having computationally efficient iterations, and they need a lot longer training epochs (250 for MEST and 500 for RigL) to regain accuracy at the cost of higher training FLOPS. 
TopKAST (\cite{Top-KAST2021}) always prunes highest magnitude weights (uniform sparsity), but updates a superset of active weights based on gradients of this superset (backward sparsity) so that the pruned out weights can be revived. For high quality results, they had to use full weight gradients (no backward sparsity) throughout the training.  
PowerPropagation (PP) (\cite{PowerProp2021})  transforms the weights as $w=v |{v}|^{\alpha-1}$, s.t. it creates a heavy-tailed distribution of trained weights. They observe that weights initialized close to 0 are likely to be pruned out, and weights are less likely to change sign. PP, when applied on TopKAST, i.e., TopKAST + PP, improves the accuracy of TopKAST.
RigL and MEST keep first layer dense while TopKAST and its variants make first and last layers dense.
Gradmax (\cite{GradMax2022}) proposed to grow network by adding more weights gradually with training epochs
in order to reduce overall training FLOPS. 
\textit{Dense-to-Sparse Training} :
GMP (\cite{GMP2017}) is a simple, magnitude-based weight pruning applied gradually throughout the training. Gradients do not flow to masked weights. \cite{state-sparsity2019}) improved the accuracy of GMP by keeping first layer dense and last layer sparsity at 80\%. DNW (\cite{DNW2019}) also uses magnitude pruning while allowing gradients to flow to masked weight via STE (\cite{STE2013}). 
DPF (\cite{DPF2020}) updates dense weights using full-gradients of sparse weights while simultaneously maintains dense model.
OptG (\cite{OptG2022}) learns both weights
and a pruning supermask in a gradient driven manner. They argued in favour of letting gradient flow
to pruned weights so that it solves the ‘independence paradox’ problem that prevents from achieving
high-accuracy sparse results. However, they achieve a given sparsity budget by increasing sparsity
according to some sigmoid schedule (sparsity is extracted only after 40 epochs). This suffers from
larger training FLOPs count. The achieved sparsity distribution is uniform across layers.
SWAT (\cite{SWAT2020}) sparsifies both weights
and activations by keeping Top magnitude elements in order to further reduce the FLOPS.

\textit{Learned Pruning: }
For a given sparsity budget, the methods discussed above either keeps sparsity uniform at each layer or heuristically creates non-uniform sparsity distribution. The following methods learns a non-uniform sparsity distribution via some sparsity-inducing optimization. 

DST (\cite{DST2020}) learns both weights and pruning thresholds (creating weight mask) where they impose exponential decay of thresholds as regularizer. A hyperparameter controls the amount of regularization, leading to a balance between sparsity and accuracy. In order to reduce the sensitivity of the hyperparameter, they manually reset the sparsity if it exceeds some predefined limit (thus occasionally falling into dense training regime). Also, approximation of gradient of pruning step function helps some masked elements to receive loss gradient.

STR (\cite{STR2020}) is a SotA method that learns weights and pruning threshold using ReLU as a mask function, where STE of ReLU is used to approximate gradients of survived weights and masked weights do not receive loss gradients. It does not use explicit sparsity-inducing regularizer. However, extracting high sparsity from early iterations leads to run-away sparsity; this forces STR to run fully dense training for many epochs before sparsity extraction kicks in. 

 SCL (\cite{SCL2022}) learns weights and a mask (same size as weights) that is binarized in forward pass. This mask along with a decaying connectivity hyperparameter are used as a sparsity-inducing regularizer in the objective function. The learned mask increases the effective model size during training, which might create overhead moving parameter from memory.

 LTP (\cite{LTP2021}) learns the pruning thresholds using soft pruning and soft L0 regularization
where sigmoid is applied on transformed weights and sparsity is controlled by a hyper-parameter. 
CS (\cite{CS2021}) uses sigmoidal soft-threshold function as a sparsity-inducing regularization. 
For a detailed discussion on related work, see \cite{sparsity-survey2021}.

\section{Gradient Annealing (GA)}

A typical pruning step of deep networks involves masking out weights that are below some threshold $T$. This sparse representation of weights benefits from sparse computation in forward pass and in computation of gradients of inputs. 
We propose the following pruning step, where $w$ is a weight and $T$ is a threshold that can be deterministic (e.g., TopK magnitude) or learnable:
\begin{eqnarray*}
\text{(sparse)} \quad \tilde w = \text{sign}(w) \cdot h_\alpha(|w| - T)
\end{eqnarray*}
\begin{eqnarray}
\label{eqn:step_function}
\text{Forward pass} \quad h_\alpha(x) = 
\begin{cases}
x, \quad x > 0 \\
0, \quad x \leq 0
\end{cases}
\qquad 
\text{(Proxy) Gradient} \quad \frac{\partial h_\alpha (x)}{\partial x} = 
\begin{cases}
1, \quad x > 0 \\
\alpha, \quad x \leq 0
\end{cases}
\end{eqnarray}

where $0 \leq \alpha \leq 1$. $\tilde w$ is $0$ if $|w|$ is below threshold $T$. Magnitude-based pruning is a greedy, temporal view of parameter importance. However, some of the pruned-out weights (in early training epochs) might turn out to be important in later epochs when a more accurate sparse pattern emerges. For this, $h_\alpha(\cdot)$ in eqn (\ref{eqn:step_function}) allows the loss gradient to flow to masked weights in order to avoid permanently pruning out some important weights. 
The proposed gradient approximation is inspired by the Straight Through Estimator (STE) \cite{STE2013} which replaces zero gradients of discrete sub-differentiable functions by proxy gradients in back-propagation. 
Furthermore, we decay this $\alpha$ as the training progresses. We call this technique the \textit{Gradient Annealing}. 

\begin{figure}[!t]
\centering
\begin{subfigure}[b]{0.46\textwidth}
\includegraphics[scale=0.45]
{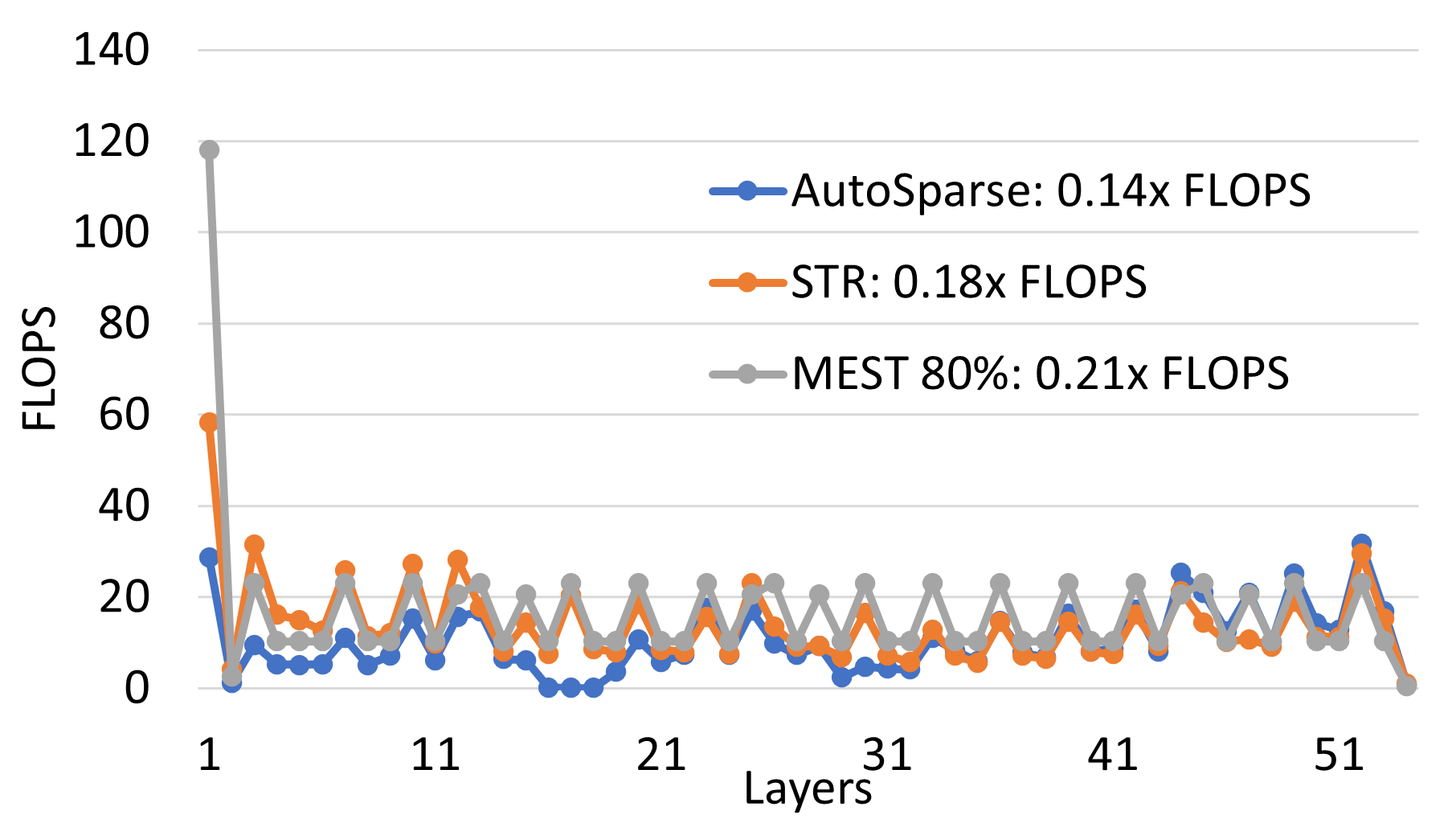}
\caption{FLOPS (M) for 80\% sparse ResNet50 produced by uniform-MEST and learned sparsity methods} 
\label{fig:compute_profile}
\end{subfigure}
\qquad 
\begin{subfigure}[b]{0.46\textwidth}
\includegraphics[scale=0.4]
{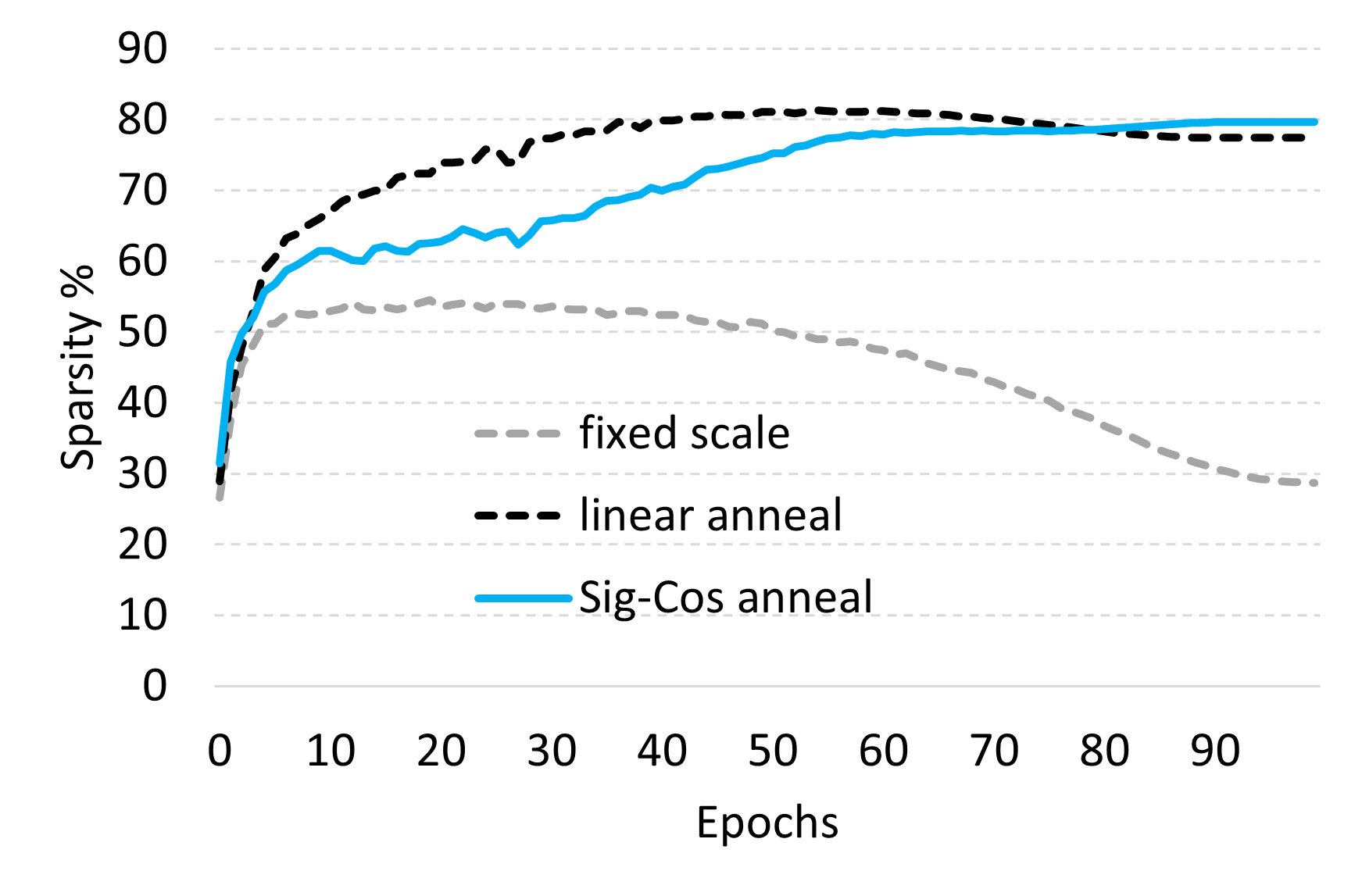}
\caption{Sparsity achieved for various gradient annealing} 
\end{subfigure}
\caption{Sparse ResNet50 training on ImageNet}
\label{fig:sparsity_profile}
\end{figure}

We decay $\alpha$ at the beginning of every epoch and keep this value fixed for all the iterations in that epoch. We want to decay  $\alpha$ slowly in early epochs and then decay it steadily. For this, we compare several choices for decaying $\alpha$: fixed scale (no decay), linear decay, cosine decay (same as learning rate (\ref{eqn:cosine_decay})), sigmoid decay (defined in (\ref{eqn:sigmoid_decay})) and Sigmoid-Cosine decay (defined in  (\ref{eqn:sig_cos_decay})).
For sigmoid decay in (\ref{eqn:sigmoid_decay}), $L_0=-6$ and $L_1=6$. 
For total epochs $T$, scale for $\alpha$ in epoch $i$ is 
\begin{eqnarray}
\text{Cosine-Decay}(i,T) \qquad c_i &=& (1 + \text{cosine}(\pi \cdot i / T))/2 \label{eqn:cosine_decay} 
\\
\text{Sigmoid-Decay}(i,T) \qquad s_i &=& 1 - \text{sigmoid}(L_0 + (L_1-L_0)\cdot i/T)\label{eqn:sigmoid_decay}
\\
\text{Sigmoid-Cosine-Decay}(i,T) \qquad &=& \max\{s_i, c_i\}
\label{eqn:sig_cos_decay}
\end{eqnarray}
Figure \ref{fig:sparsity_profile} shows the effect of various linear and non-linear annealing of $\alpha$ on dynamic sparsity. Fixed scale with no decay (STE) does not give us a good control of dynamic sparsity. Linear decay is better than this but suffers from drop in sparsity towards the end of training. Non-linear decays in eqn (\ref{eqn:cosine_decay},  \ref{eqn:sigmoid_decay}, \ref{eqn:sig_cos_decay}) provide much superior trade-off between sparsity and accuracy. 
While eqn (\ref{eqn:cosine_decay}) and eqn (\ref{eqn:sig_cos_decay}) show very similar behavior, sharp drop of eqn (\ref{eqn:sigmoid_decay}) towards the end of training push up the sparsity a bit (incurring little more drop in accuracy). 
These plots are consistent with our analysis of convergence of GA in eqn (\ref{eqn:convergence}). 
Annealing schedule of $\alpha$ closely follows learning rate decay schedule. 

\textbf{Analysis of Gradient Annealing }
Here we analyze the effect of the transformation $h_\alpha(\cdot)$ on the convergence of the learning process using a simplified example as follows.
Let $v = |w| - T$, $u = h_\alpha(v)$, optimal weights be $w^*$ and optimal threshold be $T^*$, i.e., $v^* = |w^*| - T^*$. Let us consider the loss function as
\begin{eqnarray*}
\min_{v} \mathcal{L}(v) = 0.5 \cdot (h_\alpha(v) - v^*)^2  
\end{eqnarray*}

and let $\partial h_\alpha(v)$ denote the gradient $\frac{\partial h_\alpha(v)}{\partial v }$. We consider the following cases for loss gradient for $v$. 

\begin{eqnarray}
\label{eqn:convergence}
\frac{\partial \mathcal{L}}{\partial {v}} = 
\partial h_\alpha(v)(h_\alpha(v) - v^*) 
=
\begin{cases}
\partial h_\alpha(v) \cdot 0 = 0 \quad \text{if}  \quad h_\alpha(v) = v^* \\
\partial h_\alpha(v)\cdot (v-v^*)  = 1 \cdot (v-v^*) \quad \text{if}  \quad v > 0 \text{ and } v^* > 0 \\
\partial h_\alpha(v) \cdot (v+|v^*|) = 1 \cdot (v+|v^*|) \quad \text{if}  \quad v > 0 \text{ and } v^* \leq 0 \\
\partial h_\alpha(v) \cdot (-v^*) = \alpha \cdot (-v^*) \quad \text{if}  \quad v \leq 0 \text{ and } v^* > 0 \\
\partial h_\alpha(v) \cdot (|v^*|) = \alpha \cdot (|v^*|) \quad \text{if}  \quad v \leq 0 \text{ and } v^* \leq 0 
\end{cases}
\end{eqnarray}

Correct proxy gradients for $h_\alpha(\cdot)$ should move $v$ towards $v^*$ during pruning (e.g., opposite direction of gradient for gradient descent) and stabilize it at its optima (no more updates). Therefore, $\partial h_\alpha(v) > 0$ should be satisfied for better convergence of $v$ to  $v^*$. Our $h_\alpha(\cdot)$ satisfies this condition for $\alpha >0$.
Furthermore, for $v>0$, $v$ gets updated proportional to $v-v^*$, i.e., how far $v$ is from $v^*$. As training progresses and $v$ gets closer to $v^*$, 
$v$ receives gradually smaller gradients to finally converge to $v^*$. %
However, for $v \leq 0$, $v$ receives gradient proportional to magnitude of $\alpha \cdot v^*$, irrespective of how close $v$ is to $v^*$. 
Also, note that we benefit from sparse compute when $v \leq 0$.


We set initial $T$ high in order to achieve sparsity from early epochs. However, this likely leads to a large number of weights following condition 4 in eqn (\ref{eqn:convergence}). Fixed, large $\alpha$ (close to 1) makes large correction to $v$ and moves it to $v^*$ quickly. Consequently, $v$ moves from condition 4 to condition 2, losing out the benefit of sparse compute. A lower $\alpha$ `delays' this transition and enjoys the benefits of sparsity. \textit{This is why we choose $\alpha < 1$ rather than identity STE as proxy gradient} (unlike \cite{SCL2022}). 
However, as training progresses, more and more weights move from condition 4 to condition 2 leading to a drop in sparsity. This behavior is undesirable to reach a target sparsity at the end of training. In order to overcome this, we propose to decay $\alpha$ with training epochs such that we enjoy the benefits of sparse compute while $v$ being close to $v^*$. That is, GA provides a more controlled and stable trade-off between sparsity and accuracy throughout the training.


Note that, GA is applicable when we compute loss gradients for a superset of active (non-zero) weights that take part in forward sparse computation using gradient descend. For an iteration $t$, let the forward sparsity be $S$. If $\alpha=0$, then we need to calculate gradient for only those non-zero weights as other weights would not receive gradients due to ReLU STE. In order to benefit from such computational reduction, we can set $\alpha=0$ after several epochs of $\alpha$ annealing. 

\section{AutoSparse : Sparse Training with Gradient Annealing}
AutoSparse is the sparse training algorithm that combines the best of learnable threshold pruning techniques STR with Gradient Annealing (GA). AutoSparse meets the requirements necessary for efficient training of sparse neural networks.

$\bullet$ \textbf{Learnable Pruning Thresholds} : Eliminates sorting-based threshold computation to reduce sparsification overhead compared to uniform pruning methods. Learns the non-uniform distribution of sparsity across the layers.

$\bullet$ \textbf{Sparse Model Discovery} : 
Discover an elegant trade-off between model accuracy vs. level of sparsity by applying Gradient Annealing method (as shown in Figure \ref{fig:sparsity_profile}). Produce a sparse model at the end of the training with desired sparsity level guided by the hyper-parameter $\alpha$.

$\bullet$  \textbf{Accelerate Training/Inference} :
Reduce training FLOPs by training with sparse weights from scratch, maintaining high levels of sparsity throughout the training, and using sparsity in both forward and backward pass. Produce FLOPS-efficient sparse model for inference.

Previous proposals using learnable threshold methods such as DST and STR address the first criterion but do not effectively deal with accuracy vs sparsity trade-off. These methods also do not accelerate training by reducing FLOPS as effectively as our method. Uniform pruning methods such as TopKAST, RigL, MEST address the third criterion of accelerating training, but incur sparsfication overheads for computing threshold values and cannot automatically discover the non-uniform distribution of sparsity. This leads to sub-optimal sparse models for inference (Figure \ref{fig:compute_profile}).

\textbf{Formulation: }
Let $\mathcal{D}:=\{(\bf{x}_i \in \mathbb{R}^d, \it{y_i} \in \mathbb {R})\}$ be the observed data, $\mathcal{W}$ be the learnable network parameters, $\mathcal{L}$ be a loss function. For an $L$-layer DNN, $\mathcal{W}$ is divided into layer-wise trainable parameter tensors, $[\textbf{W}_\ell]_{\ell=1}^L$. As various layers can have widely different number of parameters and also unequal sensitivity to parameter alteration, we use one trainable pruning parameter, $s_\ell$ for each layer $\ell$, i.e., $\bf{s}$ $= [s_1, ..., s_L]$ is the vector of trainable pruning parameter. Let $g: \mathbb{R}\rightarrow \mathbb{R}$ be applied element-wise. For layer $\ell$, $T_\ell = g(s_\ell)$ is the pruning threshold for $\textbf{W}_\ell$. 
We seek to optimize:
\begin{eqnarray}\label{eqn:main}
\min_{\mathcal{W},\bf{s}}\mathcal{L} (\mathcal{S}_{h_\alpha,g}(\mathcal{W},\bf{s}), \mathcal{D})
\end{eqnarray}
where, function $\mathcal{S}_{h_\alpha,g}$, parameterized by $h_\alpha \in \mathbb{R}\rightarrow \mathbb{R}$ 
that is applied element-wise. 
%
\begin{eqnarray}
\label{eqn:prune}
\hat{\textbf{W}}_\ell = \mathcal{S}_{h_\alpha,g}(\textbf{W}_\ell,s_\ell) = \text{sign}(\textbf{W}_\ell)\cdot h_\alpha(|\textbf{W}_\ell| - g(s_\ell))
\end{eqnarray}

Gradient annealing is applied via $h_\alpha(\cdot)$ as discussed earlier. 

\textbf{Sparse Forward Pass: } 
At iteration $t$, for layer $\ell$, sparse weights are $\hat{\textbf{W}}{}^{(t)}_\ell = \mathcal{S}_{h_\alpha,g}(\textbf{W}^{(t)}_\ell,s^{(t)}_\ell)$ as defined in eqn (\ref{eqn:prune}). 
Let the set of non-zero (active) weights   $\mathcal{A}^{(t)}_\ell = \{i : \hat{\textbf{W}}{}^{(t)}_{\ell,i} > 0\}$. For simplicity, let us drop the subscript notations. $\hat{\textbf{W}}$ is used in the forward pass as 
$\textbf{Y}=\textbf{X}\otimes\hat{\textbf{W}}$ where $\otimes$ is tensor MAC operation, e.g., convolution. Let $A$ denote the fraction of non-zero elements of $\textbf{W}$ belonging to $\mathcal{A}$, i.e, out forward sparsity is $1 - A$.  
$\hat{\textbf{W}}$ is also used in computation of input gradient during backward pass. Every iteration, we update ${\textbf{W}}$ and construct $\hat{\textbf{W}}$ from updated ${\textbf{W}}$ and learned threshold. 

\textbf{Sparse Compute of Input Gradient: }
For an iteration and a layer (we drop the subscripts $t$, $\ell$ for simplicity), output of sparse operations in forward pass need not be sparse, i.e., $\textbf{Y}=\textbf{X}\otimes\hat{\textbf{W}}$ is typically dense. 
Consequently, gradient of output $\nabla_Y$ is also dense. We compute gradient of input $\nabla_X$ as $\hat{\textbf{W}}\otimes \nabla_Y$. Computation of $\nabla_X$ is sparse due to sparsity in $\hat{\textbf{W}}$. 

\textbf{Sparse Weight Gradient: }
Gradient of weights $\nabla_W$ is computed as $\textbf{X}\otimes\nabla_Y$.
Note that, for forward sparsity S, $\alpha=0$ implies weight gradient sparsity S as no gradient flows to pruned weights. We can have a hyperparameter that specifies at which epoch we set $\alpha=0$, to enjoy benefits of sparse weight gradients. 
However, we need to keep $\alpha\neq 0$ for several epochs to achieve high accuracy results, 
losing the benefits of sparse weight gradient.
In order to overcome this, we propose the following for the epochs when $\alpha\neq 0$. 
%
We can make sparse $\nabla_W$ if we compute loss gradient using a subset $\mathcal{B}$ of $\textbf{W}$.
\begin{eqnarray}
\label{eqn:back_sparsity}
\mathcal{B} = \{i:\textbf{W}_i \in \text{TopK(\textbf{W}, $B$)}\}, 
\quad B \ge A
\end{eqnarray}
where $\mathcal{B}$ is a superset of $\mathcal{A}$ and TopK($\textbf{W}$,$k$) picks indices of $k$ largest magnitude elements from $\textbf{W}$. 
This constitutes our weight gradient sparsity $1 - B$ (similar to TopKAST). 
We apply gradient annealing on set $\mathcal{B} \setminus \mathcal{A}$, i.e., gradients of weights in $\mathcal{B} \setminus \mathcal{A}$ are decayed using $\alpha$. Note that, $\alpha = 0$ implies $\mathcal{B} = \mathcal{A}$.

\textbf{FLOPS Computation: } 
For a dense model with FLOPS $f^\ell_D$ and sparse version with FLOPS $f^\ell_S$ for layer $\ell$, total dense training FLOPS for a single sample is $3 \cdot \sum_\ell f^\ell_D$ and sparse training FLOPS is $3 \cdot \sum_\ell f^\ell_S$ when only sparse weight gradient is involved (AutoSparse with $\alpha=0$), otherwise, it is  $2 \cdot \sum_\ell f^\ell_S + \sum_\ell f^\ell_D$ for full-dense weight gradient (AutoSparse with $\alpha \neq 0$). Also, for AutoSparse, $f^\ell_S$ varies with iterations, so we sum it up for all layers, all iterations and all data points to get the final training FLOPS count. For explicitly set sparsity $f_B$ for weight gradients, AutoSparse training FLOPS for a sample is $2\cdot \sum_\ell f^\ell_S+\sum_\ell f_B$ (for $\alpha \neq 0$) and $2\cdot \sum_\ell f^\ell_S+\sum_\ell \max(f_B, f^\ell_S)$ (for $\alpha = 0$).
\section{Experiments}

For \textit{Vision Models},  
we show sparse training results on ImageNet-1K (\cite{imagenet}) for two popular CNN architectures: ResNet50 \cite{resnet} and and MobileNetV1 \cite{mobilenet}, to demonstrate the generalizability of our method. 
For AutoSparse training, we use SGD as the optimizer, momentum $0.875$, learning rate (max) $0.256$ using a cosine annealing with warm up of $5$ epochs. We run all the experiments for $100$ epochs using a batch size $256$. We use weight decay $\lambda=0.000030517578125$ (picked from STR \cite{STR2020}), label smoothing $0.1$, $s_0=-5$. We presented our results only using Sigmoid-Cosine decay of $\alpha$ (defined in eqn (\ref{eqn:sig_cos_decay})).

For \textit{Language Models},  
we choose Transformer models \cite{Transformer} for language translation on WMT14 English-German data. We have 6 encoder and 6 decoder layers with standard hyperparameter setting: optimizer is ADAM with betas $(0.9, 0.997)$, token size $10240$, warm up $4000$, learning rate $0.000846$ that follows inverse square root decay. 
We apply AutoSparse by introducing a learnable threshold for each linear layer, and we initialize them as $s_0=-7.0$.
Also, initial $\alpha_0=0.4$ is annealed according to exponential decay as follows. For epoch $t$ and $\beta > 0$ (we use $\beta=1$): 
\begin{eqnarray}
\label{eqn:exponential_decay}
\text{Exponential-Decay } (t; \beta) = e^{(-\beta \cdot t)}
\end{eqnarray}
 We keep the first and last layers of transformer  dense and apply AutoSparse to train it for $44$ epochs.
We repeat the experiments several times with different random seeds and report the average numbers.

\textbf{Notation: } We define the following notations that are used in the tables.
`Base': Dense baseline accuracy, `Top1(S)': Top-1 accuracy for sparse models, `Drop\%': relative drop in accuracy for sparse models from Base, `$S\%$': percentage of model sparsity, `Train F': fraction of training FLOPs comparing to baseline FLOPs, `Test F': fraction of inference FLOPs comparing to bsaeline FLOPs, `Back $S\%$': explicitly set sparsity in weight gradients ($B$ in eqn (\ref{eqn:back_sparsity})), `BLEU(S)': BLEU score for sparse models. Smaller values of `Train F' and `Test F' suggests larger reduction in computation.

\begin{table*}[!t]
\centering
    \begin{tabular}{ p{2.3cm} || p{0.8cm} | p{0.8cm}  p{0.7cm} p{0.8cm} | p{1.0cm}   p{1.0cm}  p{3.3cm}}
Method & Base & Top1(S) & Drop\% & S\% & Train F & Test F &
comment   
\\ \hline 
{}
& & & &  
& & &
\\
[-5pt]
RigL
& $76.8$ & $74.6$ & $2.86$ & $80$ 
& $0.33\times$ & $0.22\times$ & uniform sparsity 
\\\\
[-10pt] 
TopKAST$^\star$
& $76.8$ & $75.7$ & $0.94$ & $80$ 
& $0.48\times$ & $0.22\times$ & uniform sparsity  
\\\\
[-10pt] 
TopKAST$^\star$+PP
& $76.8$ & $76.24$ & $0.73$ & $80$  
& $0.48\times$ & $0.22\times$ & uniform sparsity  
\\\\
[-10pt]
TopKAST$^*$+\textbf{GA} 
& $76.8$ & $76.47$ & $0.43$ & $80$ & 
$0.48\times$ & $0.22\times$ & uniform sparsity  
\\\\
[-10pt]
MEST$_{1.7\times}$+EM 
& $76.9$ & $76.71$ & $0.25$ & $80$ & 
$0.57\times$ & $0.21\times$ & uniform sparsity  
\\\\
[-10pt] 
DST 
& $74.95$ & $74.02$ & $1.24$ & $80.4$   
& -- & $0.15\times$ & learnable sparsity
\\\\
[-10pt] 
STR 
& $77.01$ & $76.19$ & $1.06$ & $79.55$ & $0.54\times$  
& $0.18\times$ & learnable sparsity  
\\\\
[-10pt] 
\textbf{AutoSparse}
& $77.01$ & $76.77$ & \textbf{$0.31$} & $79.67$ & $0.51\times$  
& \textbf{$0.14\times$} & $\alpha_0$=.75,$\alpha$=0@epoch90 
\\\\
[-10pt]
\textbf{AutoSparse}
& $77.01$ & $76.59$ & \textbf{$0.55$} & $80.78$ & $0.46\times$  
& \textbf{$0.14\times$} & $\alpha_0$=.8,$\alpha$=0@epoch70 
\\\\
[-10pt]
\hline
\\
[-8pt]
MEST$_{1.7\times}$+EM 
& $76.9$ & $75.91$ & $1.29$ & $90$ & 
$0.28\times$ & $0.11\times$ & uniform sparsity  
\\\\
[-10pt] 
DST 
& $74.95$ & $72.78$ & $2.9$ & $90.13$ 
& -- & $0.087\times$ & learnable sparsity
\\\\
[-10pt] 
STR 
& $77.01$ & $74.31$ & $3.51$ & $90.23$ & $0.44\times$  
& $0.083\times$ & learnable sparsity  
\\\\
[-10pt] 
\textbf{AutoSparse}
& $77.01$ & $75.9$ & $1.44$ & $85.1$ & $0.42\times$  
& \textbf{$0.096\times$} & $\alpha_0$=.9,$\alpha$=0@epoch50  
\\\\
[-10pt]
\textbf{AutoSparse}
& $77.01$ & $75.19$ & $2.36$& $89.94$ & $0.40\times$  
& \textbf{$0.081\times$} & $\alpha_0$=.9,$\alpha$=0@epoch45
\\\\
[-10pt]
\hline
\\
[-8pt]
STR 
& $77.01$ & $70.4$ & $8.58$ & $95.03$ & $0.28\times$  
& $0.039\times$ & learnable sparsity  
\\\\
[-10pt] 
\textbf{AutoSparse}
& $77.01$ & $70.84$ & \textbf{$8.01$} & $95.09$ & $0.2\times$  
& \textbf{$0.036\times$} & $\alpha_0$=0.8,$\alpha$=0@epoch20 
\\
[3pt]
\end{tabular}
\caption
{ ResNet50 on ImageNet: 
Comparing accuracy, sparsity and the FLOPS (dense $1\times$) for training and inference for selected sparse training methods. \textbf{TopKAST$^\star$}: TopKAST with $0\%$ backward sparsity. \textbf{TopKAST$^{\star}$+GA}: TopKAST$^{\star}$ with Gradient Annealing boosts the accuracy despite having same training/inference FLOPS. 
For AutoSparse $79.67\%$ sparsity, $\alpha_0$=0.75 is decayed till epoch 90 and then set to 0 (implying $\sim 80\%$ forward and backward sparsity after epoch 90). Similarly, for AutoSparse $95.09\%$ sparsity, $\alpha_0$=0.8 is decayed till epoch 20 and then set to 0. Standard deviation of results for AutoSparse is less than 0.03. 
}
\label{table:flops_resnet50}
\end{table*}

\begin{table*}[!t]
\centering
    \begin{tabular}{ p{1.6cm} || p{0.8cm} | p{0.9cm}  p{0.9cm} | p{1.0cm}   p{1.0cm}  p{1.0cm} p{3.5cm}}
Method & S\%  & Top1(S) & Drop\% & Train F & Test F &
Back(S) & comment   
\\ \hline 
{}
& & & &  
& &  &
\\
[-5pt]
\textbf{AutoSparse}
& $83.74$ & $75.02$ & \textbf{$2.58$} & $0.37\times$  
& \textbf{$0.128\times$} & 50 & $\alpha_0$=1.0, $s_0$=-8, w grad 50\% sparse
\\\\
[-10pt] 
TopKAST
& $80$ & $75$ & \textbf{$2.34$} & $0.33\times$  
& \textbf{$0.22\times$} & 50 & uniform sparse  fwd \& in grad 80\% w grad 50\%  
\\
[3pt]
\end{tabular}
\caption
{ ResNet50 on ImageNet: 
AutoSparse with explicitly set sparsity for weight gradient. 
For AutoSparse $83.74\%$ sparsity, $\alpha_0=1$ is decayed till epoch 100 resulting in identical sparsity for forward and input gradients, along with 50\% sparsity for weight gradient throughout the training (involves invoking TopK method).
}
\label{table:back_sparse_resnet50}
\end{table*}

\subsection{Efficacy of Gradient Annealing}
We compare AutoSparse results with STR and DST. 
Lack of gradient flow to pruned out elements prevents STR to achieve the optimal sparsity-accuracy trade-off. For example, they need dense training for many epochs in order to achieve high accuracy results, losing out the benefits of sparse training. Our gradient annealing overcomes such problems and achieves much superior sparsity-accuracy trade-off (Table \ref{table:flops_resnet50}, \ref{table:flops_mobilenet}).
 %
 Similarly, our method achieves higher accuracy than DST for both $80\%$ and $90\%$ sparsity budget for ResNet50 (Table \ref{table:flops_resnet50}). However, DST uses separate sparsity-inducing regularizer, whereas our gradient annealing itself does the trick.
SCL reports $0.23\%$ drop in accuracy at $74\%$ sparsity for ResNet50 on ImageNet with inference FLOPs $0.21\times$ of baseline). LTP produces 89\% sparse ResNet50 that suffers from 3.2\% drop in accuracy from baseline (worse than ours).  
Continuous Sparsification (CS) induces sparse training using soft-thresholding as a regularizer. The sparse model produced by them is used to test Lottery Ticket Hypothesis (retrained). Lack of FLOPs number makes it harder to directly compare with our method. 
GDP (\cite{GDP2021}) prunes channels using FLOPS count as a regularizer to achieve optimal accuracy vs compute trade-off (0.4\% drop in accuracy with $0.49\times$ inference FLOPS). This method is not directly comparable with our results for unstructured sparsity. 
%
%
Finally, MEST (SOTA for uniform sparse training) achieves comparable accuracy for 80\% sparse Resnet50, however, using 12\% more training FLOPS and 50\% more inference FLOPS (as their sparse model is not FLOPS-efficient). 

\textbf{Results for AutoSparse with Predetermined Backward Sparsity}

In AutoSparse, the forward sparsity is determined by weights that are below the learned threshold (let this dynamic sparsity be S). We decay $\alpha=0.85$ throughout sparse training and apply TopK to select indices of 50\% largest magnitude weights (during forward pass) that are used in computation of loss gradients (S $\ge$ 50\%). These weights are superset of non-zero weights used in forward computation. This way, we have sparsity S for forward and input gradient compute, and 50\% sparsity for weight gradient. Gradient annealing is applied on gradients of those weights that appear in these top 50\% but were pruned in forward pass. We compare our results with TopKAST 80\% forward sparsity and 50\% backward sparsity in Table \ref{table:back_sparse_resnet50}.  
We produce 83.74\% sparse model that significantly reduces inference FLOPs ($0.128\times$ of baseline)  while achieving similar accuracy and training FLOPs as TopKAST although TopKAST takes $1.7\times$ more inference FLOPS than ours.

\subsection{AutoSparse for Language Models}
For \textit{Language models}, 
we applied our AutoSparse on Transformer model using exponential decay for $\alpha_0=0.4$. This annealing schedule resulted in a better trade-off between accuracy and sparsity. Also, we set $\alpha=0$ after $10$ epochs to take advantage of more sparse compute in backward pass. 
With baseline dense BLEU $27.83$, AutoSparse achieves a BLEU score of $27.57$ (0.93\% drop) with overall sparsity $60.99\%$.  
For STR ($\alpha=0$), we used weight decay $0.01$ with initial $s$=-20, and we achieve BLEU score 27.49 (drop 1.22\%) at sparsity 59.25\%.  

\begin{table*}[!t]
\centering
    \begin{tabular}{ p{1.6cm} || p{0.8cm} | p{1.0cm}  p{1.cm} p{0.8cm} | p{1.cm}   p{1.0cm}  p{3.2cm}}
Method & Base & Top1(S) & Drop\% & S\%  & Train F & Test F &
comment   
\\ \hline 
{}
& & & &  &
& &  
\\
[-5pt]
STR 
& 71.95 & $68.35$ & $5$ & $75.28$ & $0.43\times$  
& $0.18\times$ & learnable sparsity 
\\\\
[-10pt] 
\textbf{AutoSparse}
& 71.95 & $69.97$ & \textbf{$2.75$} & $74.98$ & $0.53\times$  
& \textbf{$0.21\times$} & $\alpha_0$=.4,$\alpha$=0 @ epoch 90 
\\\\
[-10pt] 
\hline
\\
[-8pt]
STR 
& 71.95 & 64.83 & \text{9.9} & 85.8 & $0.37\times$   
& \textbf{$0.1\times$} & learnable  sparsity
\\\\
[-10pt] 
\textbf{AutoSparse}
& 71.95 & $64.87$ & {9.84} & $86.36$ & $0.30\times$
& $0.1\times$ & $\alpha_0$=.8,$\alpha$=0 @ epoch 20 
\\\\
[-10pt] 
\textbf{AutoSparse}
& 71.95 & $64.18$ & \textbf{$10.8$}  & $87.72$ & $0.25\times$  
& \textbf{$0.08\times$} & $\alpha_0$=.6,$\alpha$=0 @ epoch 20  
\\
[3pt]
\end{tabular}
\caption
{MobileNetV1 on ImageNet: 
Comparing accuracy, sparsity and FLOPS (dense $1\times$) for training and inference for selected sparse training methods. AutoSparse achieves significantly higher accuracy for comparable sparsity. 
For AutoSparse $74.98\%$ sparsity, $\alpha=0.4$ is decayed till epoch 90 and then set to 0. For AutoSparse ($87.72\%$,$86.36\%$) sparsity, ($\alpha=0.6$,$\alpha=0.8$) decayed till epoch 20 and then set to 0. Setting $\alpha=0$ at epoch $t$ implies identical forward and backward sparsity after epoch $t$. Standard deviation for AutoSparse is $0.04$.
}
\label{table:flops_mobilenet}
\end{table*}

\section{Discussion}
The purpose of gradient annealing is to establish a trade-off between dynamic, learnable sparsity and (high) accuracy where we can benefit from sparse compute from early epochs. Our AutoSparse can learn a sparsity distribution that is significantly more FLOPS-efficient for inference comparing to uniform sparsity methods.    
However, at extreme sparsity (where we sacrifice accuracy) we have observed that sparse-to-sparse training methods using much longer epochs ($2.5\times$ - $5\times$) with uniform sparsity recovers loss of accuracy significantly (MEST, RigL), and can outperform learnable sparsity methods in terms training FLOPS. Learnable sparsity methods reach target sparsity in a gradual manner incurring more FLOPS count at early epochs, whereas uniform sparsity methods have less FLOPS per iterations (ignoring other overheads for longer training). 
We like to study our AutoSparse for extended training epochs, along with (a) infrequent dense gradient update, (b) sparse gradient update in order to improve accuracy, training FLOPS and memory footprint.
Also, appropriate choice of  gradient annealing may depend on the learning hyperparameters, e.g., learning rate decay, choice of optimizer, amount of gradient flow etc.
It would be interesting to automate various choices of gradient annealing across workloads. 

\begin{table*}[!t]
\centering
    \begin{tabular}{ p{1.6cm} || p{0.8cm} | p{0.8cm} | p{1.2cm}  p{1.2cm} | p{4.0cm}}
Method & BLEU & S\%  & BLEU(S) & Drop\% & comment   
\\ \hline 
{}
& & & & & 
\\
[-5pt]
STR
& $27.83$
& $59.25$ & $27.49$ & \textbf{$1.22$} & our implementation
\\\\
[-10pt] 
\textbf{AutoSparse}
& $27.83$ & $60.99$ & $27.57$ & \textbf{$0.93$} & $\alpha_0$=0.4, $\alpha = 0$ at epoch 10
\\
[3pt]
\end{tabular}
\caption
{ 
Transformer on WMT: AutoSparse achieves better accuracy at higher sparsity than STR. Standard deviation for AutoSparse BLEU and STR BLEU are $0.08$ and $0.09$, respectively.
}
\label{table:WMT}
\end{table*}

\clearpage
\bibliography{sparse_reference} \bibliographystyle{iclr2023_conference}


\clearpage
\section{Appendix A}
\subsection{Gradient Computation}
\label{sec:grad_compute}
Let $\mathbf{I}(\cdot)$ denote an indicator function, $\left<\cdot, \cdot\right>$ denote inner product, $\odot$ denote elementwise product,  and $\mathcal{S}$ denote  $\mathcal{S}_{\textit{h}_\alpha,\textit{g}}$. 
$g$ should be continuous so that we can apply gradient descend for learning. Using the definition in eqn (\ref{eqn:step_function}), loss gradients for $\mathbf{W}_\ell$ and $s_\ell$ are 
\begin{eqnarray}
\nonumber
\mathcal{L}^{(t)} 
\leftarrow
\mathcal{L}(\mathcal{S}(\mathcal{W}^{(t)},\textbf{s}^{(t)}), \mathcal{D}), 
&&  
\mathbf{G}_\ell^{(t)} \leftarrow 
\nabla_{\mathcal{S}(\mathbf{W}_\ell, s_\ell)}\mathcal{L}^{(t)}
\\
\nonumber
\nabla_{\mathbf{W}_\ell^{(t)}}\mathcal{S}(\mathbf{W}_\ell, s_\ell) 
\leftarrow
\mathbf{G}_I^{(t)}, 
&&
\mathbf{G}_I^{(t)} 
\leftarrow
\textbf{I}\{\mathcal{S}(\mathbf{W}_\ell^{(t)},s_\ell^{(t)})\neq 0 \} + \alpha \cdot 
\textbf{I}\{\mathcal{S}(\mathbf{W}_\ell^{(t)},s_\ell^{(t)})= 0 \}
\\
(\text{Loss grad for } \mathbf{W}) 
&&
\nabla_{\mathbf{W}_\ell^{(t)}}\mathcal{L}^{(t)}
\leftarrow
\mathbf{G}_\ell^{(t)} \odot 
\mathbf{G}_I^{(t)} 
\\
(\text{Loss grad for } s)
&&
\nabla_{{s}_\ell^{(t)}} \mathcal{L}^{(t)}
\leftarrow
-g'(s_\ell^{(t)}) \text{ } \left< \mathbf{G}_\ell^{(t)} , \text{sign}(\mathbf{W}_\ell^{(t)})\odot
\mathbf{G}_I^{(t)} 
\right>
\end{eqnarray}
Apart from the classification loss, standard $L_2$ regularization is added on $\mathbf{W}_\ell$ and $s_\ell$, $\forall \in [L]$ with weight decay hyperparameter $\lambda$. Gradients received by $\mathbf{W}_\ell$ and $s_\ell$ for regularization are $\lambda \cdot \mathbf{W}_\ell$ and $\lambda \cdot s_\ell$, respectively. Parameter updates involve adding momentum to gradients, and multiplying by learning rate $\eta$. 
We choose Sigmoid function for $g(s)$ as in \cite{STR2020}.

\begin{figure}[h]
\centering
\includegraphics[scale=0.5]
{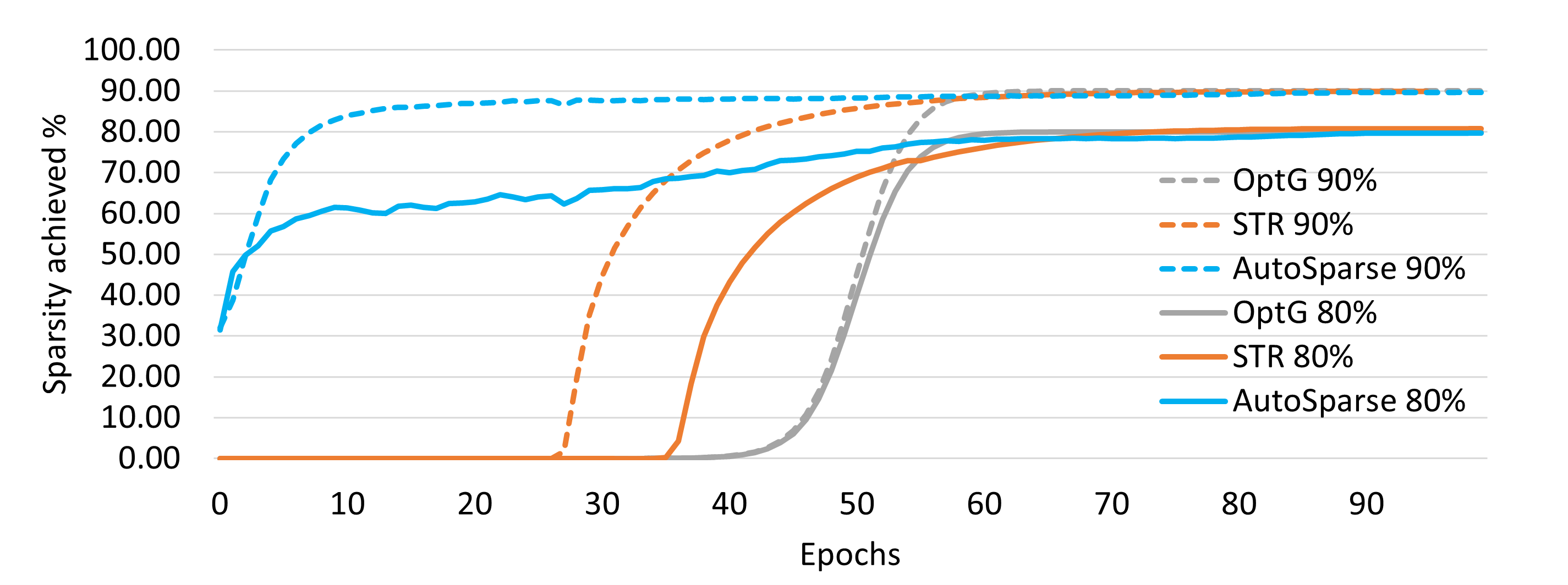}
\caption{Comparing Dynamic Model Sparsity for AutoSparse, STR and OptG.  
}
\label{fig:dynamic_sparsity}
\end{figure}

\subsection{Ablation Studies}
\begin{figure}[h]
\centering
\includegraphics[scale=0.5]
{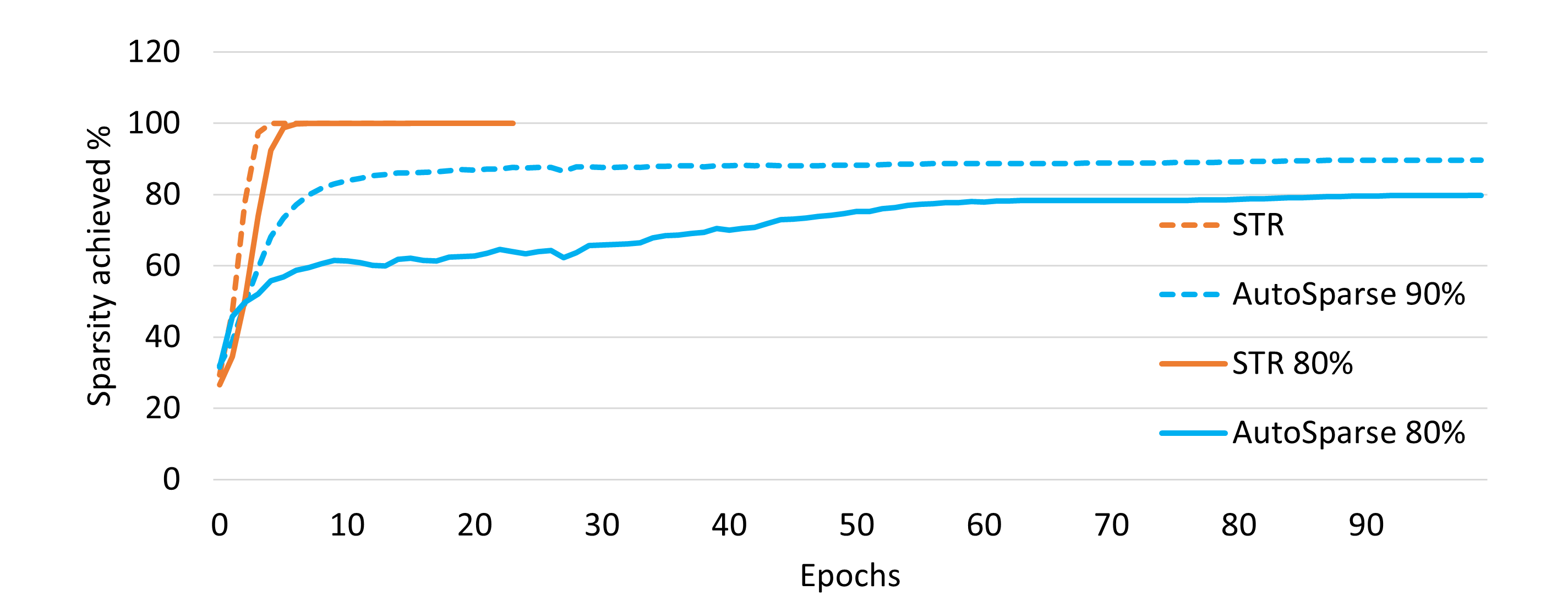}
\caption{Ablation study on early sparse training on STR and AutoSparse. $s_\text{init}=-5$ for all the cases. STR $80$: $\lambda=.000017$ (used for STR $80\%$), STR: $\lambda=.000030517578125209$, AutoSparse 80 and AutoSparse 90 both have $\lambda=0.000030517578125209$. Sparsity for both STR and STR 80 rapidly grow in an uncontrolled manner to prune out all the parameters in few epochs, while AutoSparse achieves sparsity in a controlled manner. 
} 
\label{fig:ablation_alpha}
\end{figure}

We conducted additional studies on STR (\cite{STR2020}) method to test if the average training sparsity can be improved by selecting different hyper-parameters. We initialized the threshold parameters $s_\text{init}$ with larger values to introduce sparsity in the early epochs -- we also appropriately scaled the value of $\lambda$. Figure \ref{fig:ablation_alpha} shows that when $s_\text{init}$ value is increased to $-5$ from the original value of $-3200$), the method introduces sparsity early in the training -- however the model quickly diverges after about $5$ epochs.

\subsection{Learning Sparsity Distribution}
\begin{figure}[!t]
\centering
\includegraphics[scale=0.5]
{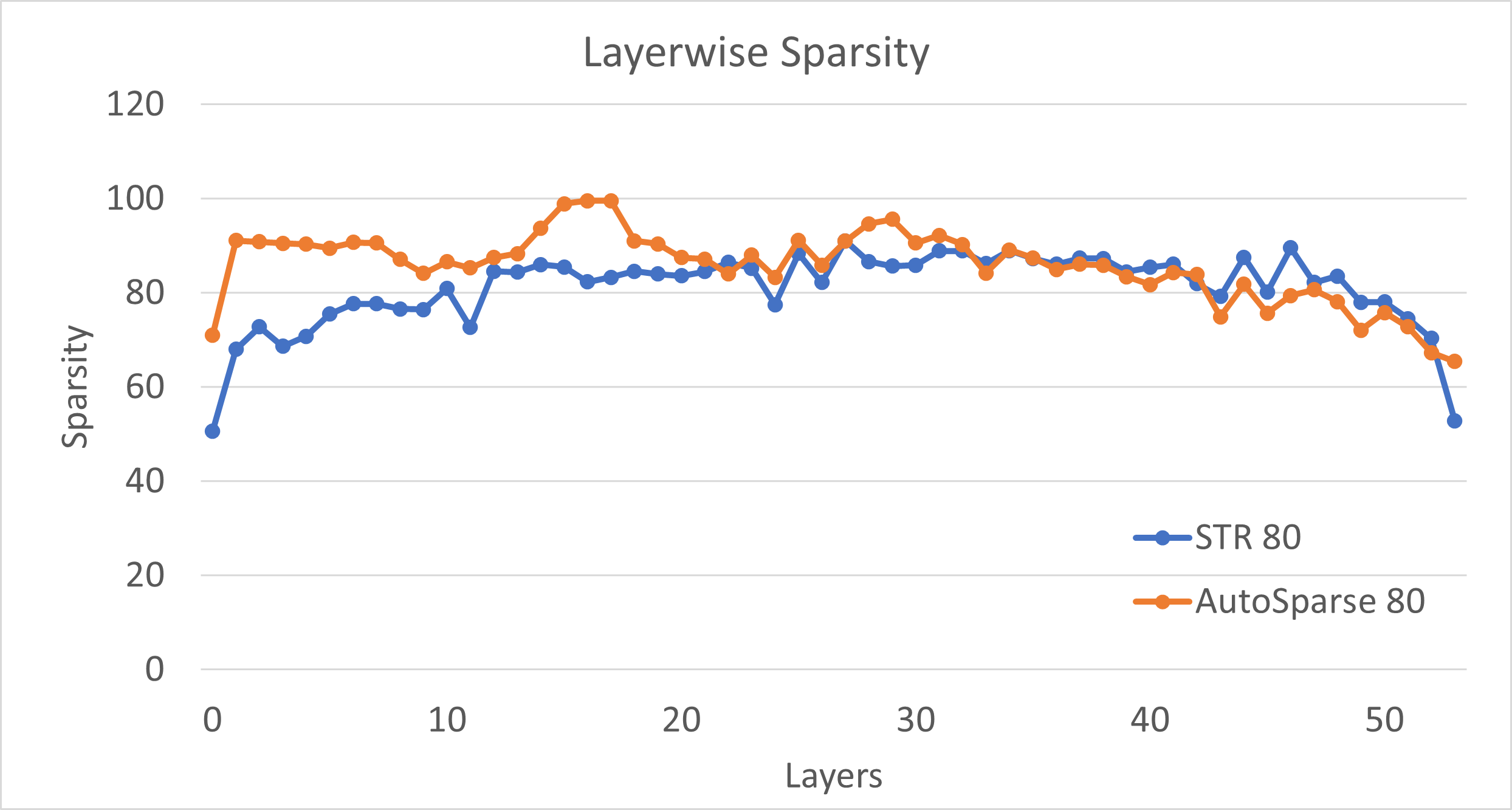}
\caption{AutoSparse learns a different sparsity distribution from STR. STR achieves higher sparsity for later layers (higher parameter density) whereas AutoSparse produces more sparsity for earlier layers, leading to significant FLOP reduction despite having the same final model sparsity. 
} 
\label{fig:sparsity_dist}
\end{figure}

AutoSparse learns a significantly different sparsity distribution than STR despite achieving similar model sparsity. AutoSparse is able to produce higher sparsity in earlier layers which leads to significant reduction in FLOPS (Figure \ref{fig:sparsity_dist}).

\subsection{Learning Sparsity Budget by Auto-Tuning $\alpha$ (AutoSparse-AT) } 
The main challenges of true sparse training are 
(1) 
the inherent sparsity (both the budget and the distribution) of various (new) models trained on various (new) data might be unknown,
(2) baseline accuracy of dense training is unknown (unless we perform dense training).
%
How can we produce a high quality sparse solution via sparse training? %
%
We seek to answer this by tuning our GA hyperparameter $\alpha$ as shown in 
%
%
Algorithm \ref{algo:autosparse}.
We have observed that increasing $\alpha$ reduces sparsity (improves accuracy) and vice versa. 
For auto-tuning of $\alpha$ (dynamically balancing sparsity and accuracy) we want the sparse loss to follow the loss trajectory of dense training. For this, we note the loss of few epochs of dense training (incurring more FLOPS count) and use it as a reference to measure the quality of our sparse training. 
For AutoSparse, we initiate the sparse training with $\alpha=0.5$ (unbiased value as $\alpha \in [0,1]$) and note the average sparse training loss at each epoch. If sparse training loss exceeds the dense loss beyond some tolerance, we increase $\alpha$ to reduce the sparsity so that the sparse loss closely follows the dense loss in the next epoch. On the other hand, when sparse loss is close to dense loss, we reduce $\alpha$ to explore more available sparsity. This tuning happens only in first few epochs of AutoSparse + AutoTune training and rest of the sparse training applies gradient annealing with the tuned value of $\alpha$.
\begin{algorithm}[!t]
\caption{AutoSparse + AutoTune}
\begin{itemize}
    \item \textbf{Input} Model $M$, data $D$, total \#epochs $T$, tuning epochs $T_0$, $\alpha_0$, reference loss $L \in \mathbb{R}^{T_0}$
    \item $\alpha \leftarrow \alpha_0$ \quad /* initialization */
    \item For $t=0,...,T-1$ epochs  
    \item \quad Train model $M$ on $D$ and calculate training loss $\mathcal{L}_t$ in (\ref{eqn:main})
    \item \quad If $t < T_0$ \qquad /* Auto-tuning phase for $\alpha$ */
    \item \qquad If $\mathcal{L}_t \ge (1+\varepsilon_0)\cdot L[t]$, \quad  $\alpha \leftarrow (1 + \varepsilon_1)\cdot \alpha$ \qquad /* reduce sparsity to control loss */
    \item \qquad Else \quad $\alpha \leftarrow (1 - \varepsilon_2) \cdot \alpha$  \qquad /* reduce $\alpha$ to explore more sparsity */
    \item \quad Else 
    \item \qquad If $t==T_0$, \quad$\alpha_0 \leftarrow \alpha$
    \item \qquad $\alpha \leftarrow \alpha_0 \cdot \text{Sigmoid-Cosine-Decay}(t-T_0, T-T_0)$ in (\ref{eqn:sig_cos_decay})
    \item \quad If $t >=90$, \quad $\alpha \leftarrow 0$ \quad /* reset $\alpha$ after certain epoch for compute benefit*/
    \item \textbf{Output} Trained sparse Model $M$
\end{itemize}
\label{algo:autosparse}
\end{algorithm}
\subsubsection{AutoSparse+Auto-Tuning Experiments}
%
We perform auto-tuning of $\alpha$ for Resnet50 and MobileNetV1 on ImageNet dataset. 
\textit{We use identical training hyper-parameters for both ResNet50 and MobileNetV1 for our AutoSparse+Auto-Tuning along with identical initial unbiased $\alpha_0=0.5$} to understand the efficacy of auto-tuning the gradient annealing. This simplifies the issues of manually-tunung many hyper-parameters.

\begin{figure}[!t]
\centering
\includegraphics[scale=0.6]
{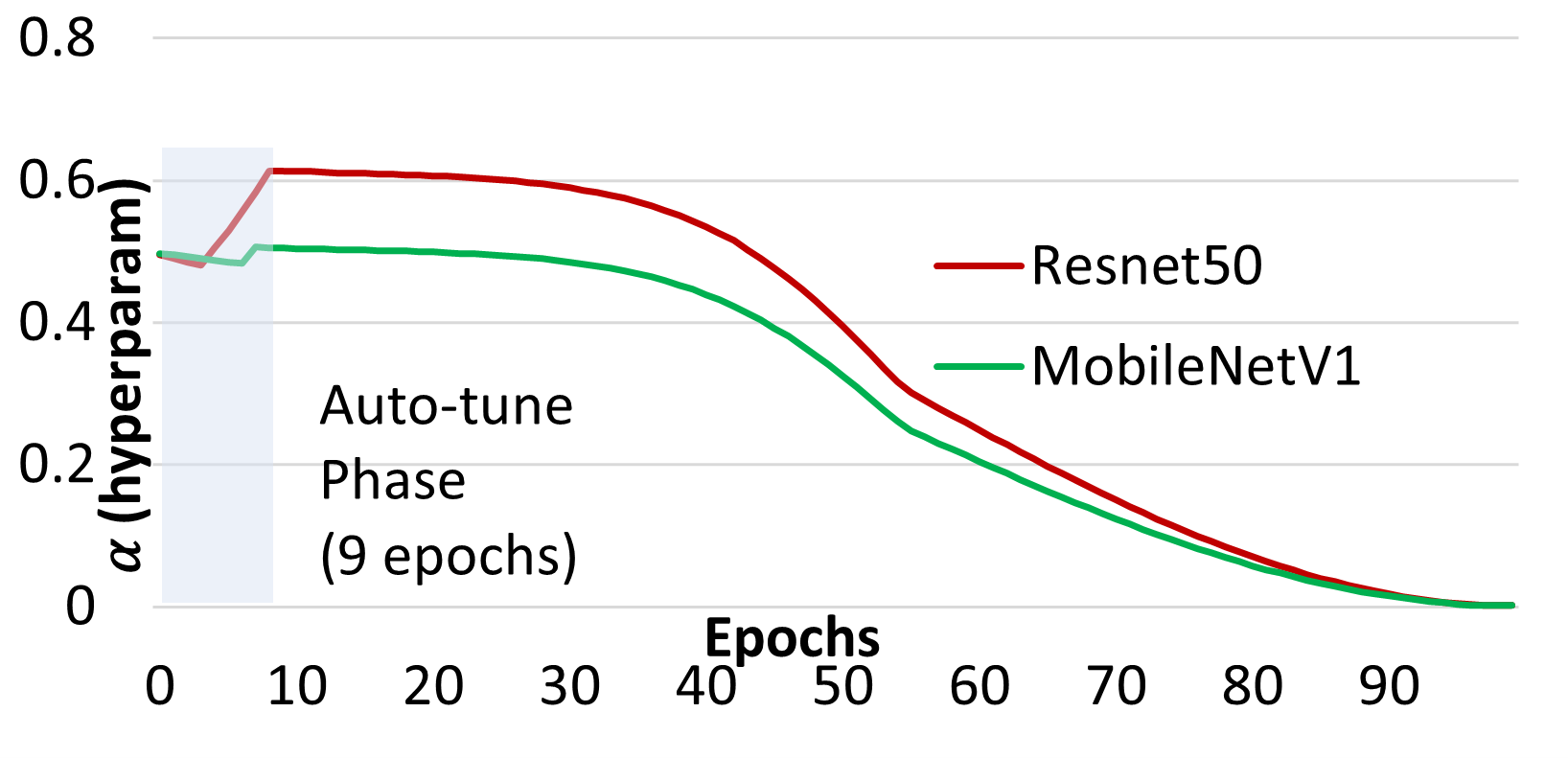}
\caption{Auto-tuning hyper-parameter $\alpha$ that dynamically controls the trade-off between accuracy and sparsity  
using average training loss of only $9$ epochs ($10\%$ budget of dense epochs) of dense training. In remaining epochs, $\alpha$ is decayed according to equation \ref{eqn:sig_cos_decay}. $\alpha_0=0.5$ irrespective of the network type.} 
\label{fig:alpha_auto}
\end{figure}
\begin{figure}[!t]
\centering
\begin{subfigure}[b]{0.47\textwidth}
\includegraphics[scale=0.5]
{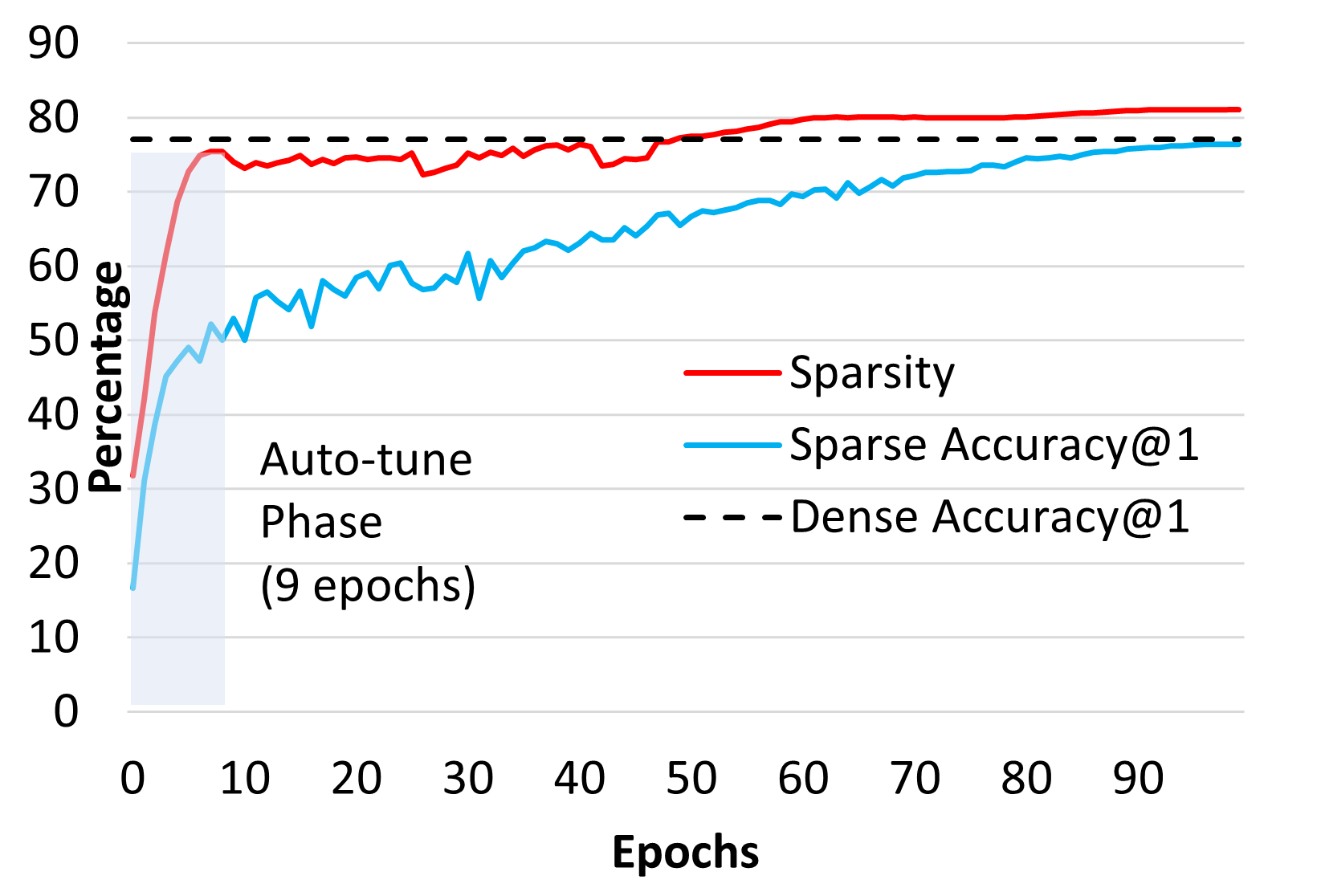}
\caption{AutoSparse+AT ResNet50} 
\label{fig:resnet50_auto}
\end{subfigure}
\qquad 
\begin{subfigure}[b]{0.47\textwidth}
\includegraphics[scale=0.5]
{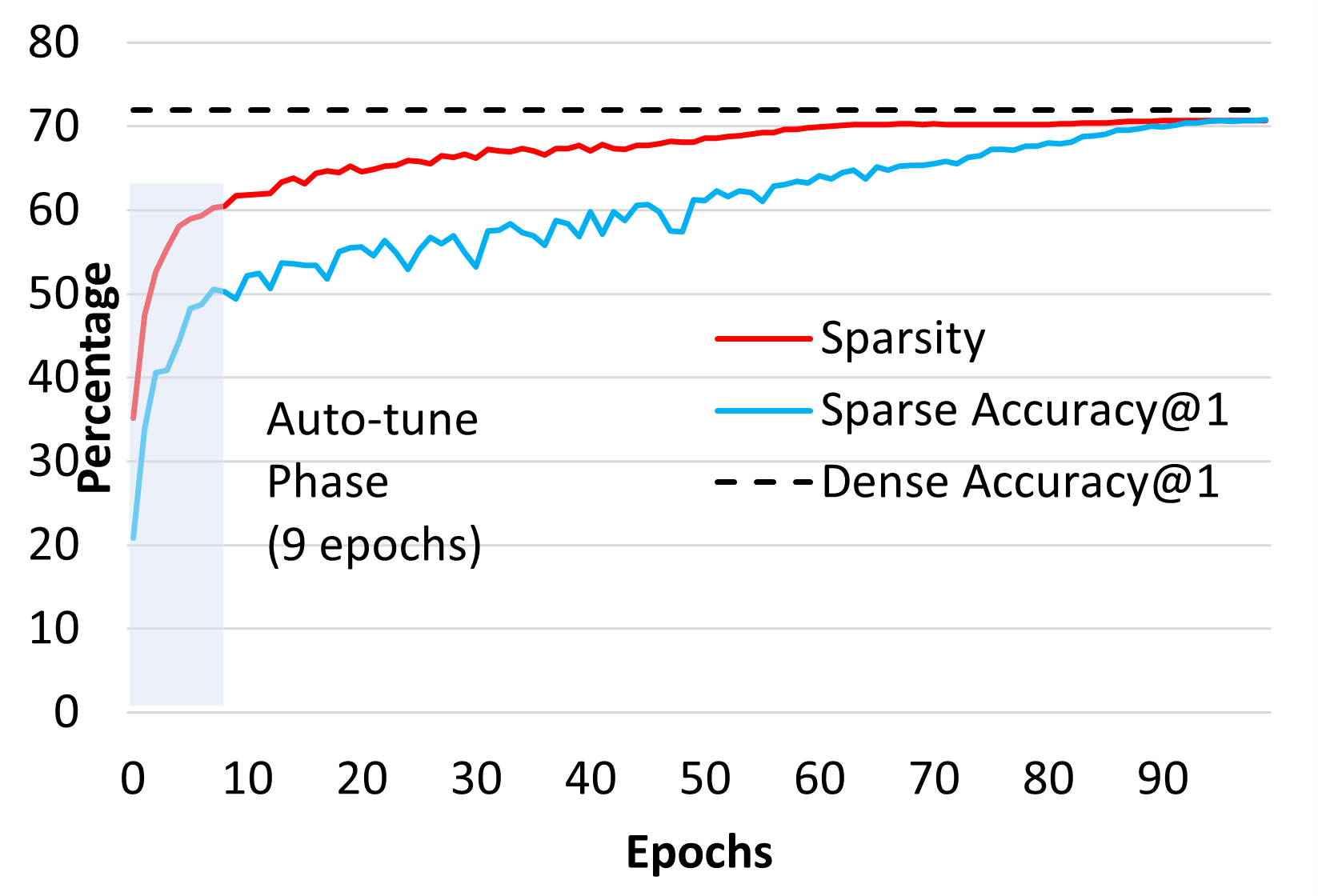}
\caption{AutoSparse+AT MobileNetV1} 
\label{fig:mobilenetv1_auto}
\end{subfigure}
\caption{AutoSparse+AT: Our auto-tuned AutoSparse almost matches dense accuracy while dynamically discovering the right amount of inherent sparsity by tuning $\alpha$ using average training loss of only $9$ epochs ($10\%$ budget of dense epochs) of dense training run.} 
\label{fig:autotune}
\end{figure}

%
%
Starting at an unbiased value $0.5$, $\alpha$ is tuned for first $9$ epochs (10\% of intended dense training epochs) of sparse training using Algorithm \ref{algo:autosparse}. 
Figure \ref{fig:alpha_auto} shows how $\alpha$ is tuned differently for ResNet50 and MobileNetV1 in a dynamic manner. This is because of the inherent sparsity budget for ResNet50 is different from that of MobileNetV1 to produce a highly accurate sparse model. 
Figures \ref{fig:resnet50_auto} and \ref{fig:mobilenetv1_auto} show how the sparsity-accuracy trade-off is discovered from early epochs for both ResNet50 and MobileNetV1. Table \ref{tab:autosparse_tuned}) summarizes various experiments.

Our AutoSparse + AutoTune can discover this in an automated manner in a single experiment. This saves a lot of training cost of manually tuning and re-running experiments to find the right balance of sparsity-accuracy.  

\begin{table}[!t]
\centering
\begin{tabular}{cc|c|c|c|c|c}
Network & Base & $(\varepsilon_0, \varepsilon_1, \varepsilon_2)$ &  
Top-1 (S) &
Sparsity & Train F & Test F \\\\[-5pt]
\hline \\[-5pt]
ResNet50 & $77.01$ & $(.01, .05, .005)$ & $76.74$ & $78.6$
& -- & --  
\\\\[-10pt] 
 & $77.01$ & $(.01, .05, .01)$ & $76.58$ & $80.1$ 
 & $0.59\times$ & $0.14\times$ 
\\\\[-10pt]
MobileNetV1 & $71.95$ & $(.01, .05, .005)$ & $70.5$ & $70.5$
& $0.67\times$ & $0.26\times$ 
\\\\[-10pt]
 & $71.95$ & $(.01, .05, .01)$ & $70.43$ & $71.2$
&--&--\\
[3pt]
\end{tabular}
\caption{
AutoSparse +AutoTune (Algorithm \ref{algo:autosparse}) accuracy vs sparsity trade-off for a budget of $100$ epochs. During auto-tuning of $\alpha$ (first $9$ epochs), we set a loss tolerance $\varepsilon_0=1\%$, increase or decrease $\alpha$ by $\varepsilon_1$ or $\varepsilon_2$, respectively. We use $\varepsilon_1=5\%$ and $\varepsilon_2$ $0.5\%$ or $1\%$. Initial $\alpha_0=0.5$ is dynamically tuned (Algorithm \ref{algo:autosparse}) and it is set to 0 at epoch 90. AutoSparse standard deviation for ResNet50 and  MobileNetV1 are $0.07$ and $0.08$, respectively.
}
\label{tab:autosparse_tuned}
\end{table}

\clearpage
\section{Appendix B}
\subsection{PyTorch Code for AutoSparse with AutoTune}
This code is built using the code base for STR \cite{STR2020}. The main changes are: (1) replace ReLU with non-saturating ReLU (class NSReLU), (2) decay the gradient scaling of negative inputs of NSReLU (neg\_grad which is denoted by $\alpha$ in the paper) every epoch, (3) auto-tune this $\alpha$ based on (dense) ref\_loss for initial few epochs (function grad\_annealing)
\\\\
\text{\#---------------------------------------------------}
\\
def set\_neg\_grad(neg\_grad\_val = 0.5):\\
\text{\quad} NSReLU.neg\_grad = neg\_grad\_val
\\\\
def get\_neg\_grad():\\
\text{\quad} return NSReLU.neg\_grad
\\\\
def set\_neg\_grad\_max(neg\_grad\_val = 0.5):\\
\text{\quad} NSReLU.neg\_grad\_max = neg\_grad\_val
\\\\
def get\_neg\_grad\_max():\\
\text{\quad} return NSReLU.neg\_grad\_max
\\
\text{\#---------------------------------------------------}
\\
class NSReLU(torch.autograd.Function):\\
\text{\quad} neg\_grad = 0.5\\
\text{\quad} neg\_grad\_max = 0.5\\
\text{\quad} \#topk = 0.5 \# for backward sparsity
\\\\
\text{\quad} @staticmethod\\
\text{\quad} def forward(self,x):\\
\text{\qquad} self.neg = x < 0\\
\text{\qquad} \#k = int(NSReLU.topk * x.numel()) \quad \# for backward sparsity\\
\text{\qquad} \#kth\_val, kth\_id = torch.kthvalue(x.view(-1), k) \quad \# for backward sparsity\\
\text{\qquad} \#self.exclude = x < kth\_val \quad \# for backward sparsity\\
\text{\qquad}  return x.clamp(min=0.0)
\\\\
\text{\quad} @staticmethod\\
\text{\quad} def backward(self,grad\_output):\\
\text{\qquad} grad\_input = grad\_output.clone()\\
\text{\qquad} grad\_input[self.neg] *= NSReLU.neg\_grad\\
\text{\qquad} \#grad\_input[self.exclude] *= 0.0 \quad \# for backward sparsity\\
\text{\qquad} return grad\_input
\\
\text{\#---------------------------------------------------}
\\
def non\_sat\_relu(x):\\
\text{\quad} return NSReLU.apply(x)
\\\\
def sparseFunction(x, s, activation=torch.relu, g=torch.sigmoid):\\
\text{\quad} return torch.sign(x)*activation(torch.abs(x)-g(s))
\\\\
def initialize\_sInit():\\
\text{\quad} if parser\_args.sInit\_type == "constant":\\
\text{\qquad} return parser\_args.sInit\_value*torch.ones([1, 1])
\\
\text{\#---------------------------------------------------}
\\
class STRConv(nn.Conv2d):\\
\text{\quad} def \_\_init\_\_(self, *args, **kwargs):\\
\text{\qquad} super().\_\_init\_\_(*args, **kwargs)
\\\\
\text{\qquad} self.activation = non\_sat\_relu
\\\\
\text{\qquad} if parser\_args.sparse\_function == `sigmoid':\\
\text{\qquad} \text{\quad} self.g = torch.sigmoid\\
\text{\qquad} \text{\quad} self.s = nn.Parameter(initialize\_sInit())\\
\text{\qquad} else:\\
\text{\qquad} \text{\quad} self.s = nn.Parameter(initialize\_sInit())
\\\\
\text{\quad} def forward(self, x):\\
\text{\qquad} sparseWeight = sparseFunction(self.weight, self.s, self.activation, self.g)\\
\text{\qquad} x = F.conv2d(
            x, sparseWeight, self.bias, self.stride, self.padding, self.dilation, self.groups
        )\\
\text{\qquad} return x
\\
\text{\#---------------------------------------------------}
\\\\
\text{\#---------------------------------------------------}
\\
from utils.conv\_type import get\_neg\_grad, set\_neg\_grad, get\_neg\_grad\_max, set\_neg\_grad\_max
\\
\text{\#---------------------------------------------------}
\\
def grad\_annealing(epoch, loss, ref\_loss):\\
\text{\quad} \# ref\_loss: reference dense loss for auto-tuning\\
\text{\quad} \# number of auto-tuning epochs\\
\text{\quad} old\_neg\_grad = get\_neg\_grad()\\
\text{\quad} new\_neg\_grad = 0.0
\\\\
\text{\quad} if epoch < len(ref\_loss):\\
\text{\qquad} dense\_loss = float(ref\_loss[str(epoch)])\\
\text{\qquad} eps\_0 = 0.01\\
\text{\qquad} eps\_1 = 0.05\\
\text{\qquad} eps\_2 = 0.005\\
\text{\qquad}  if loss > dense\_loss * (1.0 + eps\_0):\\
\text{\qquad} \text{\quad} new\_neg\_grad = old\_neg\_grad * (1.0 + eps\_1)\\ 
\text{\qquad} else:\\
\text{\qquad} \text{\quad} new\_neg\_grad = old\_neg\_grad * (1.0 - eps\_2)\\
\text{\quad} else:\\
\text{\qquad} if epoch == len(ref\_loss):\\
\text{\qquad} \text{\quad} set\_neg\_grad\_max(get\_neg\_grad())\\
\text{\qquad}  new\_neg\_grad = \_sigmoid\_cosine\_decay(\\
\text{\qquad} \text{\qquad} \text{\qquad} \text{\qquad} 
args.epochs - len(ref\_loss), epoch - len(ref\_loss), get\_neg\_grad\_max())\\
\\
\text{\quad} 
    set\_neg\_grad(new\_neg\_grad)
\\
\text{\#---------------------------------------------------}
\\    
def \_cosine\_decay(total\_epochs, epoch, neg\_grad\_max):\\
\text{\quad} PI = torch.tensor(math.pi)\\
\text{\quad} return 0.5 * neg\_grad\_max * (1 $+$ torch.cos(PI * epoch / float(total\_epochs)))
\\\\
def \_sigmoid\_decay( rem\_total\_epochs, rem\_epoch, neg\_grad\_max):\\
\text{\quad} Lmax = 6\\
\text{\quad} Lmin = -6\\
\text{\quad} return neg\_grad\_max * (1 - torch.sigmoid(torch.tensor(Lmin+(Lmax - Lmin) * \\
\text{\qquad} \text{\qquad} 
( float ( rem\_epoch ) / rem\_total\_epochs ) ) ) )
\\\\
def \_sigmoid\_cosine\_decay(rem\_total\_epochs, rem\_epoch, neg\_grad\_max):\\
\text{\quad} cosine\_scale = \_cosine\_decay(rem\_total\_epochs, rem\_epoch, get\_neg\_grad\_max())\\
\text{\quad} sigmoid\_scale = \_sigmoid\_decay(rem\_total\_epochs, \\
\text{\qquad} \text{\qquad} \text{\qquad} \text{\qquad}  
rem\_epoch, get\_neg\_grad\_max())\\
\text{\quad} return max(cosine\_scale, sigmoid\_scale)
\\\\
\text{\#---------------------------------------------------}
\\    
def train(train\_loader, model, criterion, optimizer, epoch, ref\_loss, args):\\
\text{\quad} losses = AverageMeter("Loss", ":.3f")\\
\text{\quad} top1 = AverageMeter("Acc@1", ":6.2f")\\
\text{\quad} top5 = AverageMeter("Acc@5", ":6.2f")\\
\\
\text{\quad}\# switch to train mode\\
\text{\quad} model.train()\\
\text{\quad} if epoch == 0:\\
\text{\quad} \text{\quad} set\_neg\_grad(args.init\_neg\_grad)\\
\text{\quad} batch\_size = train\_loader.batch\_size\\
\text{\quad} num\_batches = len(train\_loader)\\
\text{\quad} for i, (images, target) in tqdm.tqdm(
        enumerate(train\_loader), \\
\text{\qquad} \text{\qquad} ascii=True, total=len(train\_loader)
    ):\\
\text{\qquad} output = model(images)\\
\text{\qquad} loss = criterion(output, target.view(-1))\\
\text{\qquad} acc1, acc5 = accuracy(output, target, topk=(1,5))\\
\text{\qquad} losses.update(loss.item(), images.size(0))\\
\text{\qquad} top1.update(acc1.item(), images.size(0))\\
\text{\qquad} top5.update(acc5.item(), images.size(0))\\
\text{\qquad} optimizer.zero\_grad()\\
\text{\qquad} loss.backward()\\
\text{\qquad} optimizer.step()\\
\\
\text{\quad} \# --- anneal gradient every epoch ---
\\
\text{\quad} grad\_annealing(epoch, losses.avg, ref\_loss) \\
\text{\quad} return top1.avg, top5.avg

\newpage

\end{document}